\documentclass[a4paper]{article}

\usepackage{INTERSPEECH2022}
\usepackage[table,xcdraw]{xcolor}
\usepackage{enumitem} 

\title{Iterative pseudo-forced alignment by acoustic CTC loss \\ for self-supervised ASR domain adaptation}
\name{Fernando López$^{1, 2}$, Jordi Luque$^{1}$}
\address{
  $^1$Telefónica I+D\\
  $^2$Universidad Autónoma de Madrid}
\email{wiliam.lopezgavilanez@telefonica.com, jordi.luque@telefonica.com}

\begin{document}

\maketitle
\begin{abstract}

  High-quality data labeling from specific domains is costly and human time-consuming. In this work, we propose a  self-supervised domain adaptation method, based upon an iterative pseudo-forced alignment algorithm. The produced alignments are employed to customize an end-to-end Automatic Speech Recognition (ASR) and iteratively refined. The algorithm is fed with frame-wise character posteriors produced by a seed ASR, trained with out-of-domain data, and optimized throughout a Connectionist Temporal Classification (CTC) loss. The alignments are computed iteratively upon a corpus of broadcast TV. The process is repeated by reducing the quantity of text to be aligned or expanding the alignment window until finding the best possible audio-text alignment. The starting timestamps, or temporal anchors, are produced uniquely based on the confidence score of the last aligned utterance. This score is computed with the paths of the CTC-alignment matrix. With this methodology, no human-revised text references are required. Alignments from long audio files with low-quality transcriptions, like TV captions, are filtered out by confidence score and ready for further ASR adaptation. The obtained results, on both the Spanish RTVE2022 and CommonVoice databases, underpin the feasibility of using CTC-based systems to perform: highly accurate audio-text alignments, domain adaptation and semi-supervised training of end-to-end ASR.
\end{abstract}


\section{Introduction}

The quality and quantity of data required to train ASR systems are fundamental. Even more in end-to-end ASR systems, which require more data than hybrid Deep Neural Network (DNN) - Hidden Markov Models (HMM) to learn acoustic representations \cite{kurzinger2020ctc}. When it comes to medium and low-resource languages, the maximum exploitation of data is sought, given that manual annotations are costly and time-consuming. A common technique is to retrieve audio-to-text alignments from available audio data that has text references from the internet or TV broadcast (e.g. conferences, subtitles). Some alignment methods only work under strict conditions as human-revised transcriptions \cite{anguera2014audio}. Nevertheless, in the major part of the data available, the transcription does not exactly match the spoken content, depending on the method that has been used to generate the text (e.g. respeaking for subtitling). Therefore, other methods have explored alignment retrieval with low-quality text references \cite{hazen2006automatic, kocourautomatic, kocour2021bcn2brno, kurzinger2020ctc}. Some of these systems use a post-filtering process based on a confidence score to overcome mistaken references. Additionally, there is work with unaligned speech and text but requires word-level segmentation \cite{chen2018towards}.

More recently, the use of non-labeled data has been explored. Either in semi-supervised approaches, with pseudo-labeling of unlabeled audio \cite{synnaeve2019end}. Or with self-supervised approaches, learning speech representations that are subsequently used in a concrete task (supervised), requiring less quantity of labeled data to achieve state-of-the-art results \cite{alexei2020wav2vec2, conneau2020unsupervised, babu2021xlsr}.

In this work, we train and adapt to the TV broadcast domain an ASR in the Spanish language. Firstly, we fine-tune, only acoustically by a CTC loss, a model that was formerly trained with non-labeled data from several languages. Then, we use the trained ASR to perform utterance-level alignments on a corpus comprised of Spanish TV broadcast data. For the alignments, we developed an anchor-based algorithm that generates pseudo-forced alignments with an iterative approach. The aligned data is used to continue training the model. Hence, we propose a framework that consists of a multi-step training process that iteratively improves the former ASR performance and adapts it, step by step, to the specific broadcast TV acoustics and language domain. 

\section{Related work}

\subsection{Spanish Speech Recognition}
Recent tools and challenges have promoted the improvement of Spanish ASR. To begin with, Hugging Face's Robust Speech Recognition Challenge sought improvement in ASR systems in more than 50 languages. The participants used a fine-tuned version of Wav2Vec2-XLS-R (300M, 1B, or 2B of parameters) with the CommonVoice database, and decoding based on an n-gram language model (LM). Additionally, any pre and post-processing system, such as denoising mechanisms or spelling correctors, was allowed. The winner in the Spanish language achieved a Word Error Rate (WER) of 8.82\% in the test set of CommonVoice v6.0, reducing it to 6.27\% with an LM \cite{grosman2021xlsr-1b-spanish}. Moreover, the Speech-To-Text Albayzin Evaluation consists of automatic transcription for Spanish TV shows. The generated transcripts are compared with human revised transcriptions. In the previous editions, the Machine Learning and Language Processing (MLLP) research group achieved the best results, in 2021 a 16.06\% WER in the test split of RTVE2020DB \cite{app12020804}.

\subsection{Audio-to-Text alignments}
Audio-to-text alignments enrich transcriptions with temporal references to the spoken content. They are required in fields such as closed-captioning \cite{bordel2016probabilistic}, audiobooks or database generation for speech technologies \cite{panayotov2015librispeech, anguera2014audio}. Traditionally, alignments have been performed with HMM-based models and the Viterbi algorithm \cite{moreno1998recursive, hazen2006automatic, anguera2014audio, panayotov2015librispeech}. Frameworks as sphinx \cite{lamere2003cmu} and MAUS \cite{schiel1999automatic} are based on these technologies. This approach presents many inconveniences \cite{moreno1998recursive, kurzinger2020ctc, bordel2016probabilistic}, such as the assumption that the transcription contains the exactly spoken utterance, problems with out-of-vocabulary words, problems with long audio tracks as the Viterbi algorithm fails to scale well with speech length; and Viterbi losses may lead to a complete misaligned file due to incapacity to recover.

Recently, in \cite{kurzinger2020ctc}, a segmentation mechanism using CTC-based end-to-end networks was proposed. It shows better results for utterance-alignment in German, compared with different tools as MAUS (based in HTK \cite{young1993htk}) or Gentle (based in Kaldi \cite{povey2011kaldi}). Our work is based on this alignment mechanism so it will be described in detail in section \ref{sec:ctc-aligment}.

\section{Methodology}
\subsection{CTC-alignment algorithm}
\label{sec:ctc-aligment}
The algorithm proposed in \cite{kurzinger2020ctc} requires an end-to-end ASR trained with already aligned data and optimized with a CTC loss. Having a $T$-length acoustic observation $\bm{x}_{t}$, associated with a text reference of length $M-2$. A special token is added at the beginning of the sequence and the blank token added at the end of the sequence. Let's consider this $M$-length sequence of characters to be aligned, which indexes are represented by $j\in[1, M]$, and a discrete sequence of acoustic frames $a=\{ a_1, a_2, ... , a_T\}$ within the interval $t\in[1, T]$. The alignment task is defined as the assigning of the corresponding character index $j$ to each temporal element $a_t$ within the $a$ sequence. The alignment works as follows:
\begin{enumerate}[wide, labelwidth=0pt, labelindent=0pt]
\item The network generates frame-wise character posteriors $P(c|\bm{x}_{t})$ that are used to calculate the maximum joint probabilities ($k_{t,j}$). A trellis diagram is drawn by taking the most probable of two possible transitions: the next character in reference text or the blank symbol ($\varnothing$). The transition cost for the first character in text reference is set to zero, allowing arbitrary start of the sequence in any frame.
\item Character-wise alignments ($a_t$) are calculated with backtracking. The last character alignment, which is the start of the algorithm, is selected by taking the most probable temporal position of the character. For the rest of the alignments, the most probable transitions determine the alignments. 
\end{enumerate}

Going deeper, the joint probabilities are defined as follows:
\begin{equation}
  k_{t,j}=
    \begin{cases}
    \begin{split}
    \max(k_{t-1,j}\cdot P(\varnothing|\bm{x}_t), \\ k_{t-1,j-1}\cdot P(c_j|\bm{x}_t))
    \end{split} 
    & \text{if } t>0 \wedge j>0\\
    0,              & \text{if } t=0 \wedge j>0\\
    1,              & \text{if } t=j=0\\
    \end{cases}
  \label{eq:joint_probabilities}
\end{equation}

Considering that the most probable temporal position of character $M$ is $\psi=\arg \max_{t}{k_{t, M}}$, alignments are obtained with:
\begin{equation}
  a_t=
    \begin{cases}
    M-1, & \text{if } \psi\geq\arg \max_{t}{k_{t, M-1}}\\
    a_{t+1}, &
    \begin{split}
        \text{if }
        k_{t,a_{t+1}-1}\cdot P(c_{a_{t+1}-1}|\bm{x}_{t+1})\\
        < k_{t,a_{t+1}}\cdot P(\varnothing|\bm{x}_{t+1})
    \end{split}
    \\
    a_{t+1}-1, & \text{else }\\
    \end{cases}
  \label{eq:alignment_corrected}
\end{equation}

The algorithm produces a posterior probability ($\rho_t$) for every aligned frame, selecting the maximum probability between the aligned char or the blank token. So, we capture the spikes in the temporal $\rho_t$ sequence. Moreover, a fragment score is computed by splitting the $T$-length aligned utterance in segments of length $L$. Thus, $m_l$ is the fragment score where the index of the fragment is represented by $l\in[1, T/L]$, and is obtained by:
\begin{equation}
  m_l = \frac{1}{L} \sum_{t=lL}^{(l+1)L} \rho_t.
  \label{eq:fragment_score}
\end{equation}

Finally, the utterance alignment confidence score $s_{seg}$ is given by selecting the worst fragment score, this inflicts a penalty in mismatches between ASR results and text references. 
\begin{equation}
  s_{seg} =  \min_{l} m_l,
  \label{eq:segment_score}
\end{equation}
\subsection{Iterative pseudo-forced alignment}
The alignment method presented in section \ref{sec:ctc-aligment} cannot be used directly in our case for several reasons. As we pretend to align long audio files using a transformer-based ASR, we found the necessity to process chunks instead of the whole audio. In addition, we noticed that even human-revised transcripts present differences with the spoken content. Thus, we propose an iterative chunk processing, where several combinations of audio and text are aligned until finding the best possible match. Our proposal is implemented as follows:
%
\begin{enumerate}[wide, labelwidth=!, labelindent=0pt]
\item Pre-process audio with a Voice Activity Detector (VAD), removing only non-speech segments longer than 30 seconds. Typically opening sequences of shows and sections.
\item Split the reference text in utterances with a maximum length of 24 words, as this is the maximum word sequence in CommonVoice.
\item Calculate initial time references based on total text length and total speech time. We assign an audio duration proportional to text length. This assumes constant speech velocity but is only used to have initial time references.
\item Read audio and utterances from the last temporal anchor point. In the first step, the previous anchor point is defined by the first voice event in the audio file. As the algorithm progresses, new anchor points will be defined by the accepted alignments. 
%
\begin{figure}[b!] 
\vspace{-0.1cm}
  \centering
  \includegraphics[width=\linewidth]{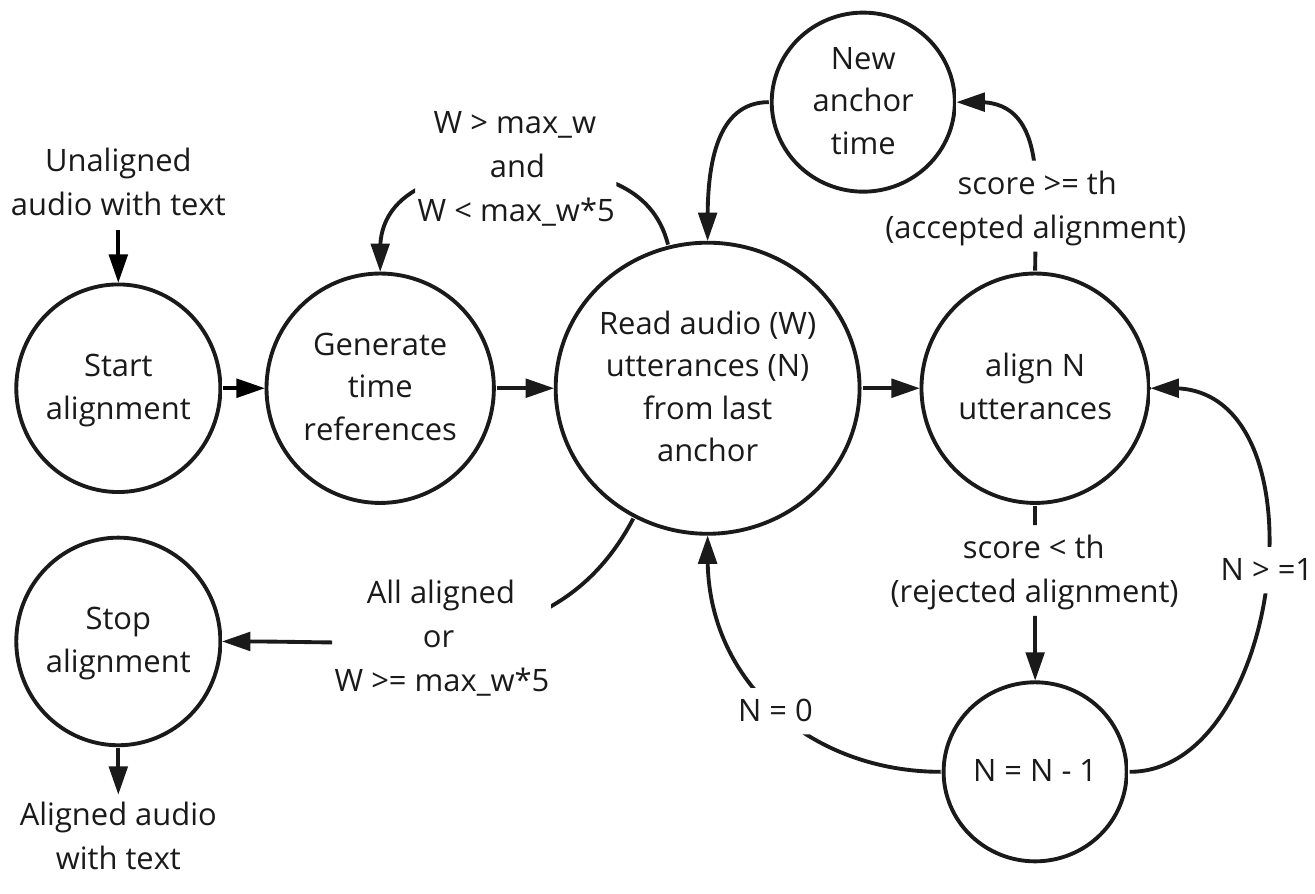}
  \caption{Simplified alignment algorithm. $th$ represents the alignment threshold, $W$ stands for the audio window length, $N$ is the number of utterances to align, and $max\_w$ stands for the maximum allowed window until recalculate time references. Iterations whenever a good alignment has been found are not considered.}
  \label{fig:state-machine}
  \vspace{-0.1cm}
\end{figure}
\item Perform iterative alignments with a fixed quantity of audio and a variable amount of text. First, get frame-wise character posteriors for the acoustic observation with the ASR. Then, perform alignments while the number of utterances is bigger or equal to one. The confidence score of the last aligned utterance is the unique criterion used to accept or reject the alignments. We use a fixed threshold, $s_{seg}=0.13$ (-2.0 in log space), to select anchor points. If the last aligned utterance provides a $s_{seg}<threshold$, we reject alignments. Therefore, we remove an utterance from the text reference and try again. In contrast, if $s_{seg}>=threshold$ we can accept the alignments, so we store them. Nevertheless, we continue iterations by removing utterances from text, looking for a better confident score. This process is repeated until finding a bad alignment or until there is no improvement. That is to say, after finding a good alignment if we find an even better alignment, then the alignment with the best confidence score is stored. If we exhaust the number of utterances without any acceptable alignment, we go back to the previous step to read more audio and text.
\end{enumerate}

\begin{figure}[t!] 
  \centering
  \includegraphics[width=\linewidth]{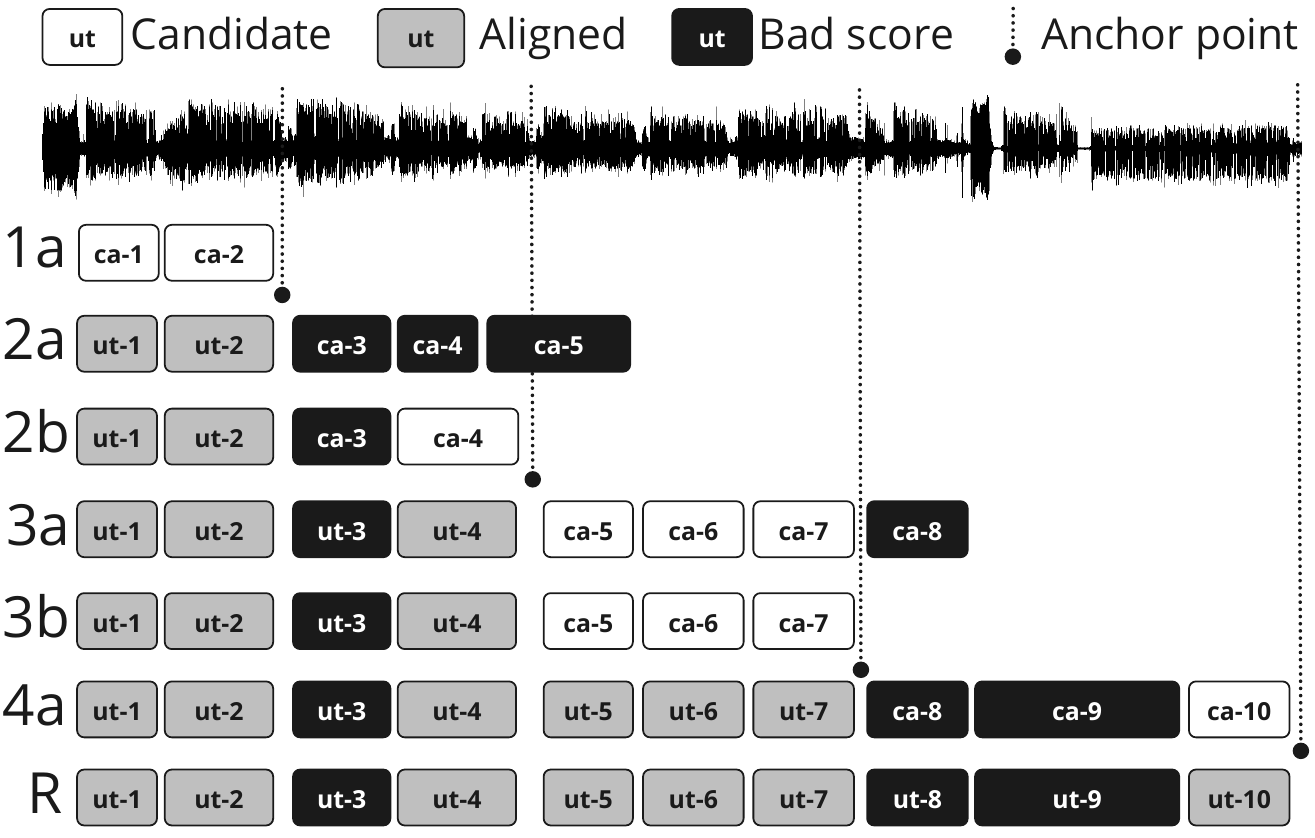}
  \vspace{-0.35cm}
  \caption{Iterations of pseudo-forced alignment algorithm. Same number represents the same audio window. Similarly, different letters represent a different set of utterances under alignment. In iterations 2b and 4a, the algorithm accepts alignments with $s_{seg}<threshold$ between anchors. $R$ is the resultant alignment, with 7 utterances with $s_{seg}$ above the alignment threshold and 3 utterances under the alignment threshold.}
  \label{fig:pseudo-forced}
  \vspace{-0.3cm}
\end{figure}

In addition, we do not allow short utterances to be anchors. Concretely, utterances with lengths inferior to or equal to $L=30$ frames (0.6 seconds) cannot be anchors. We modify the obtained $s_{seg}$ by decreasing its value. This is done because of the inability of the algorithm to provide a bad alignment confident score for short utterances. A simplified version of the algorithm is presented in Figure \ref{fig:state-machine}. Furthermore, as our unique criterion to accept the alignments of a set of utterances is the confidence score of the last aligned utterance, the algorithm allows bad alignments between anchors, thus, we call it pseudo-forced. This is especially useful for low-quality transcriptions. Figure \ref{fig:pseudo-forced} shows an example of six iterations of the algorithm.
\section{Experimental setup}
\subsection{Databases}
The proposal of this work is assessed with CommonVoice \cite{ardila2019common} as out-of-domain database and RTVE2022 \cite{RTVE2022database} as target domain database.
Spanish CommonVoice dataset comprises more than 200 hours of reading speech, which is utterance-level aligned and validated by volunteers \cite{ardila2019common}. Table \ref{tab:commonVoice-database} presents the amount of data of the two versions used in this work.

\begin{table}[t]
  \caption{Number of hours and sentences for the different splits of the Spanish CommonVoice datasets.}
  \vspace{-0.1cm}
  \label{tab:commonVoice-database}
  \centering
  \begin{tabular}{*{7}{c}}
    \toprule
    \textbf{version} & \multicolumn{2}{c}{\textbf{train}} & \multicolumn{2}{c}{\textbf{dev}} & \multicolumn{2}{c}{\textbf{test}} \\
         & hours & N & hours & N & hours & N \\
    \midrule
    6.1     & 236.9 & 161.8k & 25.2 & 15.1k & 25.9 & 15.1k \\
    7.0     & 291.2 & 196.0k & 25.8 & 15.3k & 26.4 & 15.3k  \\
    \bottomrule
  \end{tabular}
  \vspace{-0.2cm}
\end{table}

RTVE2022DB combines the databases used in previous evaluations: 2018 and 2020 with released data in 2022 \cite{RTVE2022database}. RTVE2018DB is a collection of shows from public Spanish Television (RTVE) during the years 2015 to 2018. It has 569 hours of unaligned audio, partitioned into 4 different subsets: train, dev1, dev2, and test. The train split consists of 460 hours of audio with closed captions from TV shows. The dev1, dev2, and test splits contain 57, 15, and 41 hours of human-revised transcripts, respectively. Whereas RTVE2020DB is a collection of shows from RTVE during the years 2019 to 2020. It includes a test split with 55 hours of human transcribed audio.

\subsection{End-to-end ASR}
\label{sec:end-to-end-asr}
A purely end-to-end acoustic model has been used. Based on the large XLS-R architecture, which consist of a convolutional feature encoder, followed by a transformer with 24 blocks with an inner dimension of 4096 and a model dimension of 1024. Concretely, we used the XLSR-53 model, which was pre-trained with 56k hours of audio in 53 languages \cite{conneau2020unsupervised}, and then fine-tuned in the Spanish CommonVoice dataset (version 6.1). We added two linear layers randomly initialized to the original Wav2Vec2 architecture. The resultant model counts for more than 300M trainable parameters. Finally, the transcription is obtained using simple Greedy decoding from the frame-wise character posteriors that the model produces.

Two versions of the model were built. Firstly, a model that classifies 87 characters without punctuation, as a result of retrieving any unique char from the CommonVoice database. Secondly, as the Spanish language has fewer characters, we removed other languages' utterances, characters, e.g., from the Cyrillic alphabet. Characters coming from different alphabets used in proper nouns are also discarded. As a result, the model classifies among 38 characters, including unaccented letters between a-z, the accented vowels á, é, í, ó, ú, and the dieresis on the vowel u(ü). The unique difference between both models is the number of output neurons. 

\subsection{Baseline}
The 87-char model was trained during 150 epochs with CommonVoice 6.1. While the 38-char model was trained during 15 epochs with CommonVoice 7.0. For the second model, we used the trained Wav2Vec2 weights of the first model. In the test split of the CommonVoice 7.0, the 87-char model achieved 8.8\% WER, while the 38-char model achieved 8.17\% WER. These checkpoints are our baseline experiments. Only the 38-char model was adapted to the TV domain, so, hereinafter it will be simply referred as model or ASR.

\subsection{Training setup}
Based on the CTC recipe from the SpeechBrain toolkit \cite{speechbrain}, utterance audio length is limited from 4 to 10 seconds. Similarly, signals are ordered by length, using shorter ones in the first batches of the epoch, and longer clips at the end. Regarding data augmentation, the unique technique used is SpecAugment \cite{park2019specaugment}. Furthermore, the model is optimized minimizing only a CTC loss and the learning rates (LR) are updated using the New Bob scheduler. Additionally, we manually restarted the LR at some points in training. Finally, the best checkpoint in terms of WER is stored. The ASR has been trained using a batch size of 3, setting the starting LRs for the linear layers and Wav2Vec2 of 1.0 and $10^{-5}$, respectively.

\subsection{Evaluation}
Aiming to evaluate the usefulness of our iterative pseudo-forced alignment method, we partially aligned the RTVE2018 corpus. Afterward, we adapted an out-of-domain baseline ASR by fine-tuning it through the aligned data. From this baseline model, we continued training it by combining out-of-domain data with the aligned RTVE domain data. Note that the development split comprises uniquely RTVE2018. Training and development splits of RTVE2018 have been separated by show, avoiding speaker repetition. Concretely, the development split comprises four of the total seventeen shows available.

Performance is evaluated with WER. For CommonVoice, it is computed with the SpeechBrain toolkit. For RTVE2020, transcription is generated by processing the audio file with a sliding window without overlap, and WER is computed with the sclite tool of the NIST Speech Recognition Scoring Toolkit.

\section{Results}
\subsection{Filtering methods and data recovery}
We evaluated three different methods to filter alignments. First, discarding utterances with $s_{seg}<-1.0$ (in log space). Secondly, using the Chebyshev's inequality for filtering out the 15\% worst aligned utterances. Lastly, normalizing the confidence score by considering the utterance length and then discard alignments with $s_{seg}<-1.5$ (in log space). We heuristically selected a reference length in seconds ($U=8.0$) that we consider enough to detect mismatches between reference text and ASR. Likewise, $S$ is the aligned segment length in seconds. Thus, the normalized segment score is calculated as follows: 
\begin{equation}
  s'_{seg} = \frac{s_{seg} \cdot S}{U}
  \label{eq:score_normalization}
\end{equation}

\begin{figure}[h!] 
\vspace{-0.2cm}
  \centering
  \includegraphics[width=\linewidth]{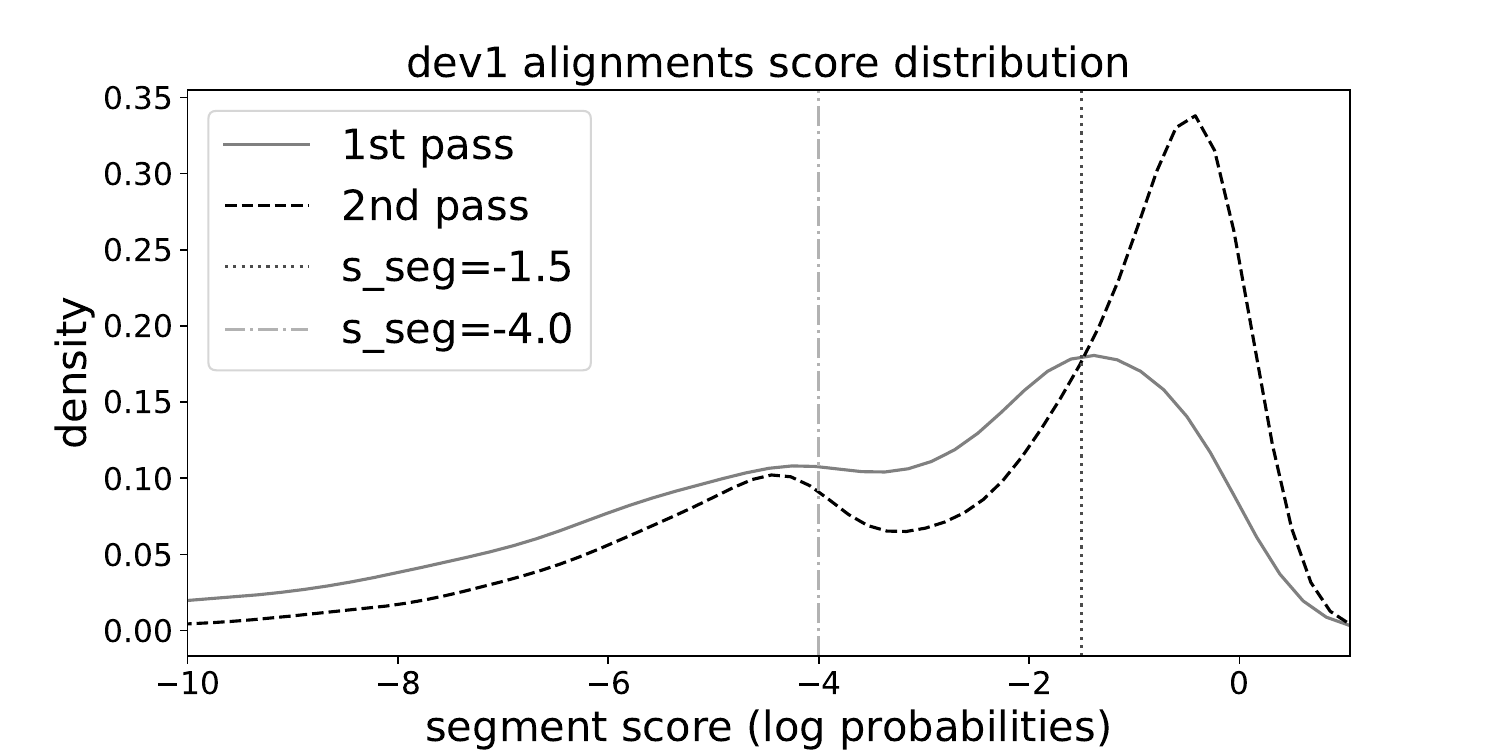}
  \vspace{-0.6cm}
  \caption{Alignment segment score ($s_{seg}$) of dev1 split from RTVE2018 across alignments. In the second pass, the confidence increases notoriously. In $s_{seg}=-4.0$, a small peak is generated due to the penalty inflicted in short utterances.}
  \label{fig:dev1-score-distribution}
  \vspace{-0.3cm}
\end{figure}

The best results are obtained while using the first method commented above, so we used only that filtering method. In Table \ref{tab:data-recovery} the quantity of aligned and filtered data is presented. The first pass is performed using the baseline ASR. For the second pass, the model is already trained with domain data (23 hours) during 15 epochs. Similarly, Figure \ref{fig:dev1-score-distribution} depicts $s_{seg}$ distribution between the two different pass alignments, showing the improvement as the model is adapted. 
\begin{table}[th]
  \caption{Data recovery, in hours, from RTVE2018DB.}
  \vspace{-0.2cm}
  \label{tab:data-recovery}
  \centering
  \begin{tabular}{*{6}{c}}
    \toprule
    \textbf{Split} & \textbf{Hours} & \multicolumn{2}{c}{\textbf{1st pass}} & \multicolumn{2}{c}{\textbf{2nd pass}}\\
    & & \text{Aligned} & \text{Filtered} &
    \text{Aligned} & \text{Filtered}\\
    \midrule
    train   & 460   & 0   & 0   & 17.5  & 14.7\\
    dev1    & 55    & 45    & 18    & 53.6  & 29.9\\
    dev2    & 15    & 8.5   & 5   & 14.3  & 9.1\\
    test    & 39    & 0   & 0   & 33.1  & 20.9\\
    \bottomrule
  \end{tabular}
\end{table}
\subsection{Domain adaptation}
\begin{figure}[h!] 
\vspace{-0.35cm}
  \centering
  \includegraphics[width=\linewidth]{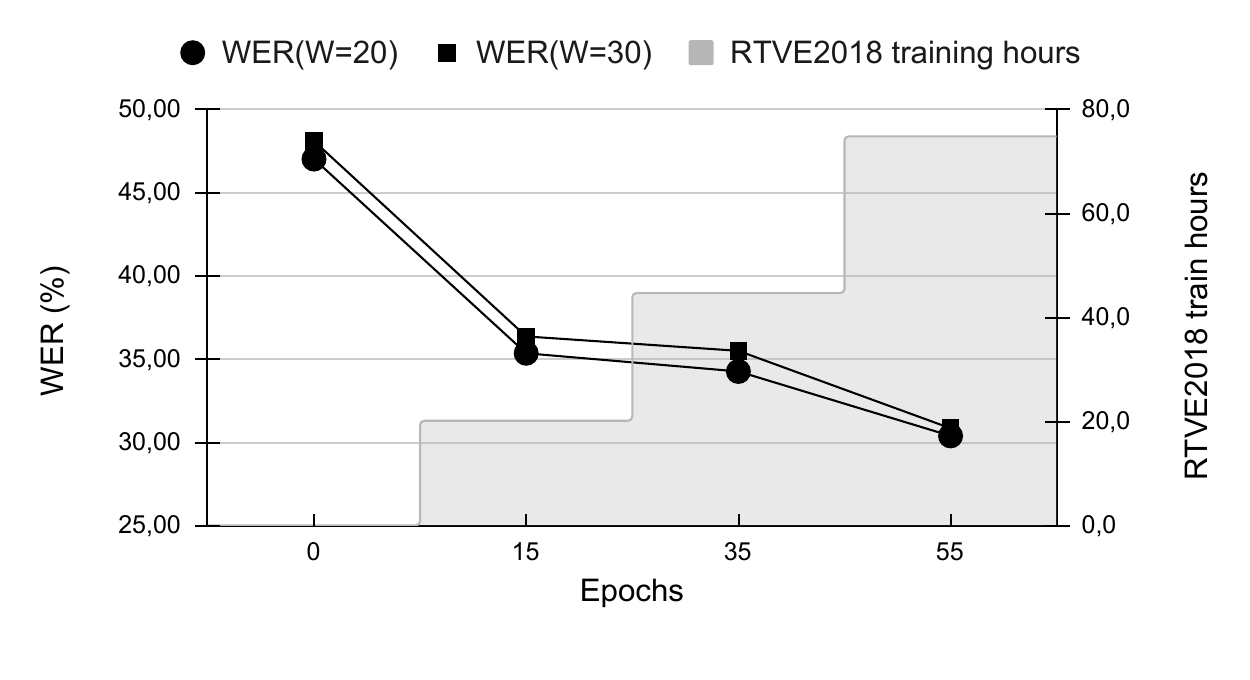}
  \vspace{-1.0cm}
  \caption{WER across training epochs. In epoch 35, LR was restarted to 1.0 for the linear layers and $10^{-5}$ for Wav2Vec2. W stands for the length in seconds of the segmentation window used to get transcriptions.}
  \label{fig:domain_adaptation}
  \vspace{-0.35cm}
\end{figure}

The domain adaptation process is presented in Figure \ref{fig:domain_adaptation} and final results are presented in Table \ref{tab:results}. Aligned data definitely contributes to the process. Best results are obtained while using the biggest amount of domain data. In RTVE2020 test, we reduced WER from 46.98\% to 30.37\%. Similarly, in CommonVoice 7.0 we reduced from 8.17\% to 7.83\%. This suggests that the use of more data, even if it is from a different domain, contributes to ASR performance in both used databases. To conclude, it is reasonable to think that there is still a margin for improvement by using more complex decoding, using an LM, or aligning more data, as only the 20\% of RTVE2018 has been aligned.
\begin{table}[h!]
  \caption{Baseline and final WER \%. DA stands for domain adaptation.}
  \vspace{-0.1cm}
  \label{tab:results}
  \centering
  \begin{tabular}{llll}
    \toprule
    \textbf{test benchmark} & \textbf{87-chars} & \textbf{38-chars} & \textbf{DA}  \\
    \midrule
    RTVE2020DB      & 47.43  & 46.98  & \textbf{30.37}  \\
    CommonVoice 6.1 & 8.51   & 8.13  & \textbf{7.80}  \\
    CommonVoice 7.0 & 8.80   & 8.17  & \textbf{7.83}  \\
    \bottomrule
  \end{tabular}
\end{table}
\section{Conclusions}
We propose an iterative pseudo-forced alignment algorithm to produce audio-to-text alignments using the CTC paths. This allowed us to perform self-supervised domain adaptation on ASR by aligning a Spanish speech corpora based on broadcast TV. Our model is a purely acoustic end-to-end model that uses simple Greedy decoding. Without any LM, we improved ASR performance over generic and domain data. Our model achieved a 35.4\% of relative WER reduction from the baseline in TV domain data. Similarly, we achieved a 4.2\% relative WER reduction in out-of-domain data. This shows the potential of the alignment method, even with low-quality text references.

\clearpage
\bibliographystyle{IEEEtran}

\end{document}